# Global Wheat Head Dataset 2021: more diversity to improve the benchmarking of wheat head localization methods.

Etienne David[1,2] *, Mario Serouart[1,2], Daniel Smith,[3] Simon Madec[1,3], Kaaviya Velumani[2,4], Shouyang Liu[5], Xu Wang[6], Francisco Pinto[7], Shahameh Shafiee[8], Izzat S. A. Tahir[9], Hisashi Tsujimoto[10], Shuhei Nasuda[11], Bangyou Zheng[12], Norbert Kichgessner[13], Helge Aasen[13], Andreas Hund[13], Pouria Sadhegi-Tehran[14], Koichi Nagasawa[15], Goro Ishikawa[16], Sébastien Dandrifosse[17], Alexis Carlier[17], Benjamin Dumont[18], Benoit Mercatoris[17], Byron Evers[6], Ken Kuroki[19], Haozhou Wang[19], Masanori Ishii[19], Minhajul A. Badhon[20], Curtis Pozniak[21], David Shaner LeBauer[22], Morten Lillemo[8], Jesse Poland[6], Scott Chapman[3,12], Benoit de Solan[1], Frédéric Baret[2], Ian Stavness[20], Wei Guo[19]

*Corresponding author. Email: etienne.david@inrae.fr

1- Arvalis, Institut du végétal, 3 Rue Joseph et Marie Hackin, 75116 Paris, France
2- UMR1114 EMMAH, INRAE, Centre PACA, Bâtiment Climat, Domaine Saint-Paul, 228 Route de l'Aérodrome, CS 40509, 84914 Avignon Cedex, France
3- School of Food and Agricultural Sciences, The University of Queensland, Gatton, 4343 QLD, Australia
4- Hiphen SAS, 120 rue Jean Dausset, Agroparc, Bâtiment Technicité, 84140 Avignon, France
5- Plant Phenomics Research Center, Nanjing Agricultural University, Nanjing, China
6- Wheat Genetics Resource Center, Dep. of Plant Pathology, Kansas State Univ., 4024 Throckmorton Plant Sciences Center, Manhattan, Kansas, United States of America
7- Global Wheat Program, International Maize and Wheat Improvement Centre (CIMMYT), Mexico, D.F., Mexico
8- Norwegian University of Life Sciences, Faculty of Biosciences, P.O. Box 5003, NO-1432 Ås, Norway
9- Agricultural Research Corporation, Wheat Research Program, P.O. Box 126, Wad Medani, Sudan
10- Arid Land Research Center, Tottori University, Tottori 680-0001, Japan
11- Laboratories of Plant Genetics and Plant Breeding, Graduate School of Agriculture, Kyoto University, Japan
12- CSIRO Agriculture and Food, Queensland Biosciences Precinct, 306 Carmody Road, St Lucia, 4067 QLD, Australia
13- Institute of Agricultural Sciences, ETH Zurich, Universitätstrasse 2, 8092 Zurich, Switzerland
14- Plant Sciences Department, Rothamsted Research, Harpenden, United Kingdom
15- Institute of Crop Science, National Agriculture and Food Research Organization, Japan
16- Hokkaido Agricultural Research Center, National Agriculture and Food Research Organization, Japan
17- Biosystems Dynamics and Exchanges, TERRA Teaching and Research Center, Gembloux Agro-Bio Tech, University of Liège, 5030 Gembloux, Belgium
18- Plant Sciences, TERRA Teaching and Research Center, Gembloux Agro-Bio Tech, University of Liège, 5030 Gembloux, Belgium
19- Graduate School of Agricultural and Life Sciences, The University of Tokyo, 1-1-1 Midori-cho, Nishitokyo City, Tokyo, Japan
20- Department of Computer Science, University of Saskatchewan, Canada

21- Department of Plant Sciences, University of Saskatchewan, Canada
22- College of Agriculture and Life Sciences, University of Arizona, Tucson, Arizona, United States of America

**Abstract**

The Global Wheat Head Detection (GWHD) dataset was created in 2020 and has assembled 193,634 labelled wheat heads from 4,700 RGB images acquired from various acquisition platforms and 7 countries/institutions. With an associated competition hosted in Kaggle, GWHD has successfully attracted attention from both the computer vision and agricultural science communities. From this first experience in 2020, a few avenues for improvements have been identified, especially from the perspective of data size, head diversity and label reliability. To address these issues, the 2020 dataset has been reexamined, relabeled, and augmented by adding 1,722 images from 5 additional countries, allowing for 81,553 additional wheat heads to be added. We now release a new version of the Global Wheat Head Detection (GWHD) dataset in 2021, which is bigger, more diverse, and less noisy than the 2020 version. The GWHD 2021 is now publicly available at http://www.global-wheat.com/ and a new data challenge has been organized on AIcrowd to make use of this updated dataset.

# 1. Introduction

Quality training data is essential for the deployment of deep learning (DL) techniques to get a general model that can scale on all the possible cases. Increasing dataset size, diversity, and quality is expected to be more efficient than increasing network complexity and depth [1]. Datasets like ImageNet [2] for classification or MS COCO [3] for instance detection are crucial for researchers to develop and rigorously benchmark new DL methods. Similarly, the importance of getting plant or crop-specific datasets is recognized within the plant phenotyping community [4]–[9], [10, p. 2], [11]–[13]. These datasets allow benchmarking the algorithm performances used to estimate phenotyping traits while encouraging computer vision experts to further improve them [10, p. 2], [14]–[17]. The emergence of affordable RGB cameras and platforms, including UAVs and smartphones, makes in-field image acquisition easily accessible. These high-throughput methods are progressively replacing manual measurement of important traits such as the counting of wheat heads. Wheat is a crop grown worldwide and the number of heads per unit area is the main component of yield potential. Creating a robust deep learning model performing for all situations requires a dataset of images covering a wide range of genotypes, sowing density and pattern, plant state and stage, as well as acquisition conditions. To answer this need for a large and diverse wheat head dataset with consistent and quality labeling, we developed the Global Wheat Head Detection (GWHD) [18] that was used to benchmark methods proposed in the computer vision community.

The GWHD dataset 2020 results from the harmonization of several datasets coming from nine different institutions across seven countries and three continents. There are already 13 publications (accessed May 2021) that have reported their wheat head detection model using GWHD dataset 2020 as the standard for training/testing data. A "Global Wheat Detection" competition hosted by Kaggle was also organized, attracting 2245 teams across the world [14], leading to improvements in wheat head detection models [19]–[22]. However, issues with the GWHD_2020 dataset were detected during the competition, including labeling noise and an unbalanced test dataset.

To provide a better benchmark dataset for the community, the GWHD_2021 dataset was organized with the following improvements: (1) the GWHD_2020 dataset was checked again to eliminate few poor-quality images; (2) images were re-labeled to avoid consistency issues; (3) a wider range of developmental stages from the GWHD2020 sites was included; (4) datasets from 5 new countries (USA, Mexico, Republic of Sudan, Norway, Belgium) were added. The resulting GWHD_2021 dataset contains 275,187 wheat heads from 16 institutions distributed across 12 countries.

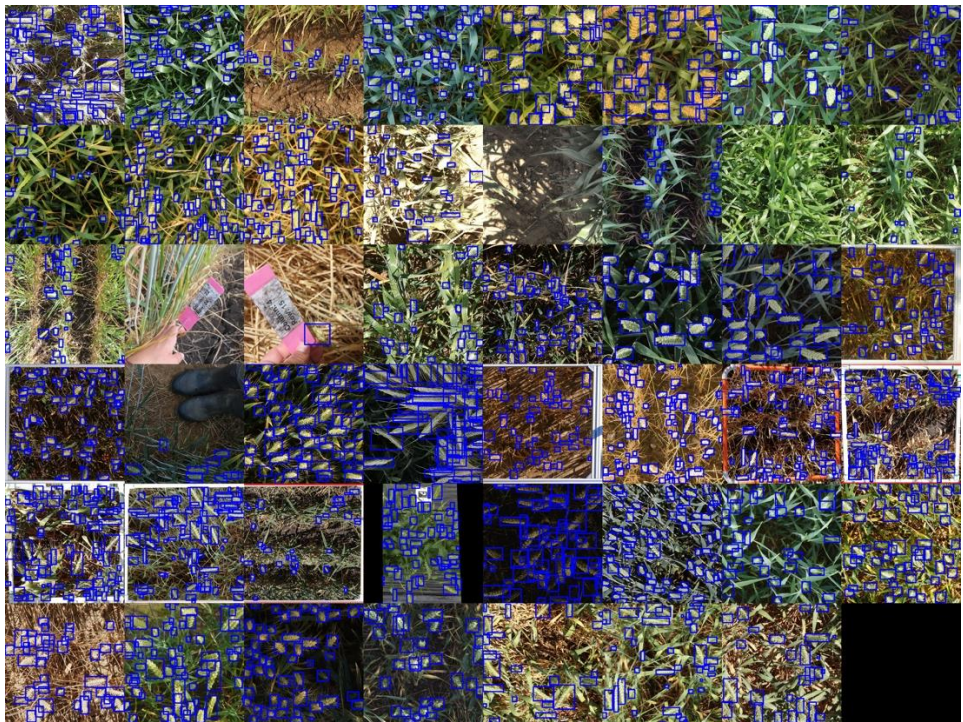

*Figure 1: Sample images of the Global Wheat Head Dataset 2021*

## 2. Materials and Methods

The first version of the Global Wheat Head Dataset published in 2020 and used for the Kaggle competition, was divided into several sub-datasets. Each sub-dataset represented all images from one location, acquired with one sensor while mixing several stages. However, wheat head detection models may be sensitive to the developmental stage: at the beginning of head emergence, part of the head is barely visible because still not fully out from the last leaf sheath, and possibly masked by the awns. Further, during ripening, wheat heads tend to bend and overlap, leading to more erratic labeling. A redefinition of the sub-dataset was hence necessary to help the user to investigate the effect of the developmental stage on model performances. The new definition of a sub-dataset was then formulated as "a consistent set of images acquired over the same experimental unit, during the same acquisition session with the same vector and sensor". A sub-dataset defines therefore a domain. This new definition forced to split the original GWHD_2020 sub-datasets into several smaller ones. The UQ_1 was split into 6 much smaller sub-datasets, Arvalis_1 was split into 3 sub-datasets, Arvalis_3 into 2 sub-datasets, Utokyo_1 into 2 sub-datasets. However, in the case of Utokyo_2 which was a collection of images taken by farmers at different stages and in different fields, the original sub-dataset was kept. Overall, the 11 original sub-datasets in GWHD 2020 were distributed into 19 sub-datasets for GWHD_2021.

Table 1 : Presentation of the sub-datasets. The column "2020 name" indicates the name given to the subdatasets for the Global Wheat Head Dataset 2020, which have been split into several new sub-datasets.

| GWHD_2021 sub-dataset name | GWHD_2020 sub-dataset name | owner | country | location | Acquisition date | Platform | Development stage | Number of images | Number of wheat head |
|---|---|---|---|---|---|---|---|---|---|
| Ethz_1 | ethz_1 | ETHZ | Switzerland | Usask | 06/06/2018 | Spidercam | Filling | 747 | 49603 |
| Rres_1 | rres_1 | Rothamsted | UK | Rothamsted | 13/07/2015 | Gantry | Filling - Ripening | 432 | 19210 |
| ULiège-GxABT_1 | | Uliège/Gembloux | Belgium | Gembloux | 28/07/2020 | Cart | Ripening | 30 | 1847 |
| NMBU_1 | | NMBU | Norway | NMBU | 24/07/2020 | Cart | Filling | 82 | 7345 |
| NMBU_2 | | NMBU | Norway | NMBU | 07/08/2020 | Cart | Ripening | 98 | 5211 |
| Arvalis_1 | arvalis_1 | Arvalis | France | Gréoux | 02/06/2018 | handheld | Post-flowering | 66 | 2935 |
| Arvalis_2 | arvalis_1 | Arvalis | France | Gréoux | 16/06/2018 | handheld | Filling | 401 | 21003 |
| Arvalis_3 | arvalis_1 | Arvalis | France | Gréoux | 07/2018 | handheld | Filling - Ripening | 588 | 21893 |
| Arvalis_4 | arvalis_2 | Arvalis | France | Gréoux | 27/05/2019 | handheld | Filling | 204 | 4270 |
| Arvalis_5 | arvalis_3 | Arvalis | France | VLB* | 06/06/2019 | handheld | Filling | 448 | 8180 |
| Arvalis_6 | arvalis_3 | Arvalis | France | VSC* | 26/06/2019 | handheld | Filling - Ripening | 160 | 8698 |
| Arvalis_7 | | Arvalis | France | VLB* | 06/2019 | handheld | Filling - Ripening | 24 | 1247 |
| Arvalis_8 | | Arvalis | France | VLB* | 06/2019 | handheld | Filling - Ripening | 20 | 1062 |
| Arvalis_9 | | Arvalis | France | VLB* | 06/2020 | handheld | Ripening | 32 | 1894 |
| Arvalis_10 | | Arvalis | France | Mons | 10/06/2020 | handheld | Filling | 60 | 1563 |
| Arvalis_11 | | Arvalis | France | VLB* | 18/06/2020 | handheld | Filling | 60 | 2818 |
| Arvalis_12 | | Arvalis | France | Gréoux | 15/06/2020 | handheld | Filling | 29 | 1277 |
| Inrae_1 | inrae_1 | INRAe | France | Toulouse | 28/05/2019 | handheld | Filling - Ripening | 176 | 3634 |
| Usask_1 | usask_1 | USaskatchewan | Canada | Saskatchewan | 06/06/2018 | Tractor | Filling - Ripening | 200 | 5985 |
| KSU_1 | | Kansas State university | US | KSU | 19/05/2016 | Tractor | Post-flowering | 100 | 6435 |
| KSU_2 | | Kansas State university | US | KSU | 12/05/2017 | Tractor | Post-flowering | 100 | 5302 |
| KSU_3 | | Kansas State university | US | KSU | 25/05/2017 | Tractor | Filling | 95 | 5217 |
| KSU_4 | | Kansas State university | US | KSU | 25/05/2017 | Tractor | Ripening | 60 | 3285 |
| Terraref_1 | | TERRA-REF project | US | Maricopa, AZ | 02/04/2020 | Gantry | Ripening | 144 | 3360 |
| Terraref_2 | | TERRA-REF project | US | Maricopa, AZ | 20/03/2020 | Gantry | Filling | 106 | 1274 |
| CIMMYT_1 | | CIMMYT | Mexico | Ciudad Obregon | 24/03/2020 | Cart | Post-flowering | 69 | 2843 |
| CIMMYT_2 | | CIMMYT | Mexico | Ciudad Obregon | 19/03/2020 | Cart | Post-flowering | 77 | 2771 |
| CIMMYT_3 | | CIMMYT | Mexico | Ciudad Obregon | 23/03/2020 | Cart | Post-flowering | 60 | 1561 |
| Utokyo_1 | utokyo_1 | UTokyo | Japan | NARO-Tsukuba | 22/05/2018 | Cart ** | Ripening | 538 | 14185 |
| Utokyo_2 | utokyo_1 | UTokyo | Japan | NARO-Tsukuba | 22/05/2018 | Cart** | Ripening | 456 | 13010 |
| Utokyo_3 | utokyo_2 | UTokyo | Japan | NARO-Hokkaido | Multi-years*** | handheld | multiple | 120 | 3085 |
| Ukyoto_1 | | UKyoto | Japan | Kyoto | 30/04/2020 | handheld | Post-Flowering | 60 | 2670 |
| NAU_1 | NAU_1 | NAU | China | Baima | n.a | handheld | Post-flowering | 20 | 1240 |

| | | | | | | | | |
|---|---|---|---|---|---|---|---|---|
| NAU_2 | | NAU | China | Baima | 02/05/2020 | cart | Post-flowering | 100 | 4918 |
| NAU_3 | | NAU | China | Baima | 09/05/2020 | cart | Filling | 100 | 4596 |
| UQ_1 | uq_1 | UQueensland | Australia | Gatton | 12/08/2015 | Tractor | Post-flowering | 22 | 640 |
| UQ_2 | uq_1 | UQueensland | Australia | Gatton | 08/09/2015 | Tractor | Post-flowering | 16 | 39 |
| UQ_3 | uq_1 | UQueensland | Australia | Gatton | 15/09/2015 | Tractor | Filling | 14 | 297 |
| UQ_4 | uq_1 | UQueensland | Australia | Gatton | 01/10/2015 | Tractor | Filling | 30 | 1039 |
| UQ_5 | uq_1 | UQueensland | Australia | Gatton | 09/10/2015 | Tractor | Filling - Ripening | 30 | 3680 |
| UQ_6 | uq_1 | UQueensland | Australia | Gatton | 14/10/2015 | Tractor | Filling - Ripening | 30 | 1147 |
| UQ_7 | | UQueensland | Australia | Gatton | 06/10/2020 | handheld | Ripening | 17 | 1335 |
| UQ_8 | | UQueensland | Australia | McAllister | 09/10/2020 | handheld | Ripening | 41 | 4835 |
| UQ_9 | | UQueensland | Australia | Brookstead | 16/10/2020 | handheld | Filling - Ripening | 33 | 2886 |
| UQ_10 | | UQueensland | Australia | Gatton | 22/09/2020 | handheld | Filling - Ripening | 53 | 8629 |
| UQ_11 | | UQueensland | Australia | Gatton | 31/08/2020 | handheld | Post-flowering | 42 | 4345 |
| ARC_1 | | ARC | Sudan | Wad Medani | 03/2021 | handheld | Filling | 30 | 888 |
| | | | | | | | Total | 6515 | 275187 |

\* VLB: Villiers le Bâcle, VSC : Villers-Saint-Christophe
\*\*Utokyo_1 and utokyo_2 were taken at the same location with different sensors
\*\*\*utokyo_3 is a special sub-dataset made from images coming large variety of farmers in Hokaido between 2016 and 2019

Almost 2000 new images were added to the Global Wheat Head Dataset, constituting a major improvement. Part of the new images come from the institutions already contributing to GWHD_2020 and were collected during a different year and/or at a different location. This was the case for Arvalis (Arvalis_7 to Arvalis_12), University of Queensland (UQ_7 to UQ_11), Nanjing Agricultural University (NAU_2 and NAU_3) and University of Kyoto (Ukyoto_1). In addition, 14 new sub-datasets were included, coming from 5 new countries: Norway (NMBU), Belgium (Université of Liège [23]), United States of America (Kansas State University [24], TERRA-REF [7]), Mexico (CIMMYT), and Republic of Sudan (Agricultural Research Council). All these images were acquired at a ground sampling distance between 0.2 and 0.4mm, i.e., similar to that of the images in the GWHD_2020. Because none of them was already labeled, a sample was selected by taking no more than one image per microplot, which was randomly cropped to 1024 x 1024px patches that we will call images in the following for the sake of simplicity.

With the addition of 1722 images and 86.000 wheat heads, the GWHD_2021 dataset contains 6500 images and 275.000 wheat heads. The increase in the number of sub-datasets from 18 to 47 leads to a larger diversity between them. However, the new definition of a sub-dataset led also to more unbalanced sub-datasets: the smallest (Arvalis_8) contains only 20 images, while the biggest (ETHZ_1) contains 747 images. This provides the opportunity to possibly take advantage of the data distribution to improve model training. Each sub-dataset has been visually assigned to several development stage classes depending on the respective color of leaves and heads (Figure 2): Post-Flowering, Filling, Filling-Ripening and Ripening. Examples of the different stages are presented in Figure 2. While being approximative, this metadata is expected to better train the models.

3. **Dataset diversity analysis**

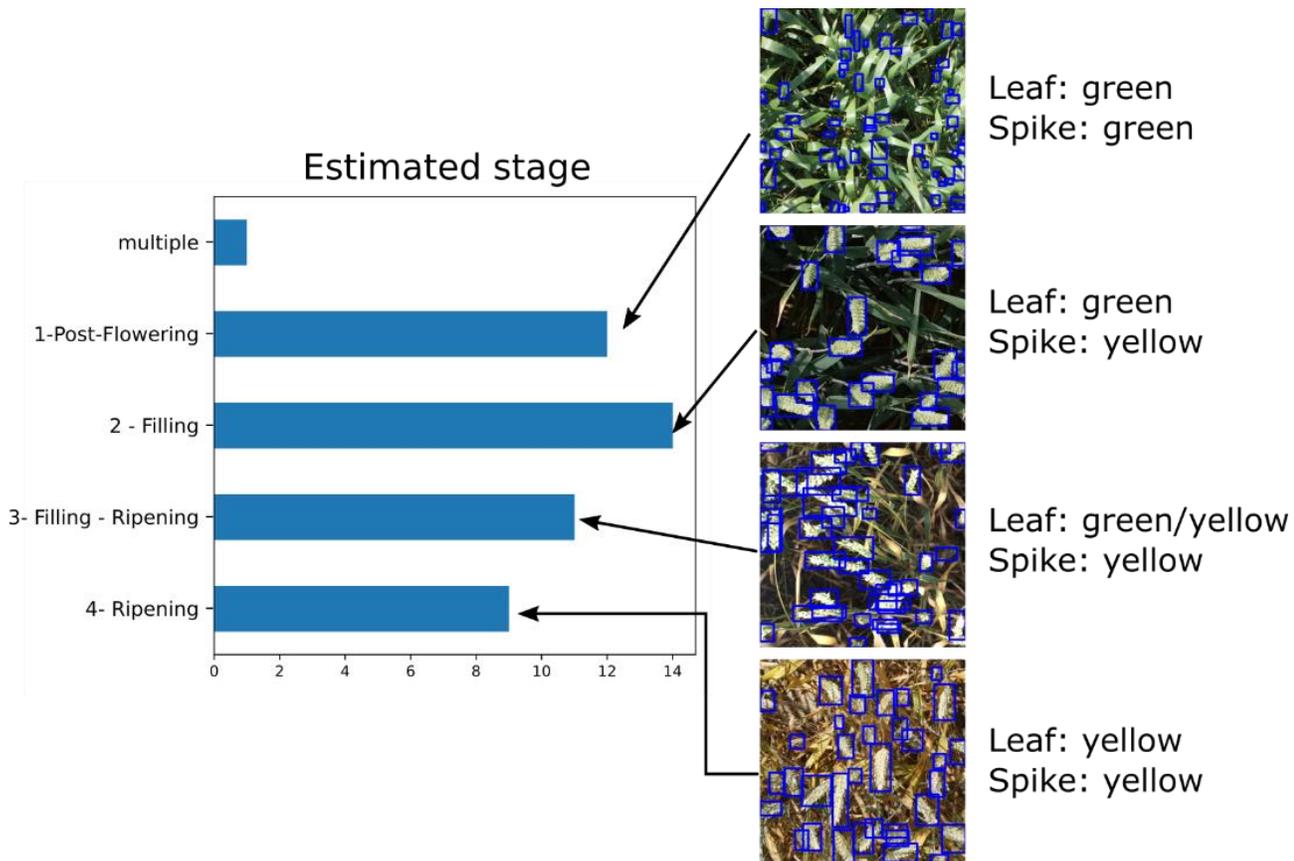

*Figure 2:Distribution of the development stage. X-axis presents the number of sub-dataset per development stage.*

In comparison to GWHD 2020, the GWHD_2021 dataset put emphasis on metadata documentation of the different subdatasets, as described in the discussion section of David et al. [18]. Alongisde the acquisition platform, each subdataset has been reviewed and a development stage has been assigned to each, except for UTokyo_3 (formerly Utokyo_2) as it is a collection of images from various farmer fields and development stages. Globally, the GWHD covers well all development stages ranging from post-anthesis to ripening (Figure 2).

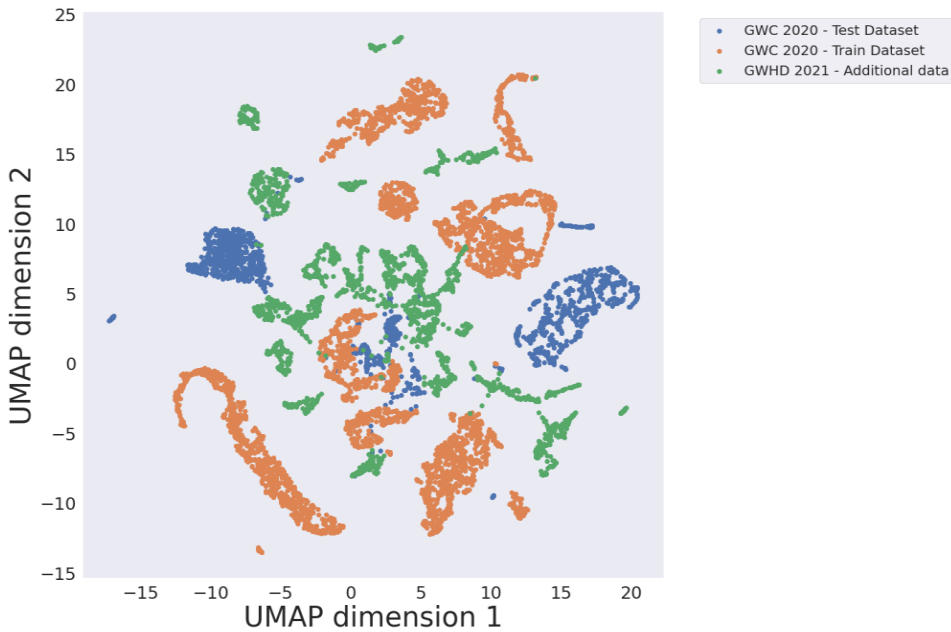

*Figure 3: Distribution of the images in the two first dimensions defined by the UMAP algorithm for the GWHD_2021 dataset. The additional sub-datasets as well as the training and test datasets from GWHD_2020 are represented by colors.*

The diversity between images within the GWHD_2021 dataset was documented using the method proposed by Tolias et al. [25]. The deep learning image features were first extracted from the VGG-16 deep network pretrained on the ImageNet dataset that is considered representating well general features of RGB images. We then selected the last layer which has a size of 14x14x512 and summed it into a unique vector of 512 channels, which is then normalized. Then, the UMAP dimentionality reduction algorithm [26] was used to project representations into a 2D space. The UMAP algorithm is used to keep the existing clusters during the projection to a low dimension space. This 2D space is expected to capture the main features of the images. Results (Figure 3) demonstrate that the test dataset used for GWHD was biased in comparison to the training dataset. The sub-datasets added in 2021 populate more evenly the 2D space which is expected to improve the robustness of the models.

4. **Presentation of Global Wheat Challenge 2021 (GWC_2021)**

The results from the Kaggle Challenge based on GWHD_2020 have been analyzed by the authors [14]. Findings emphasize that the design of a competition is critical to enable solutions that improve the robustness of the wheat head detection models. The Kaggle competition was based on a metric that was averaged across all test images, without distinction for the sub-datasets, and it was biased toward a strict match of the labelling. This artificially enhances the influence on the global score of the largest datasets such as utokyo_1 (now split into utokyo_1 and utokyo_2). Further, the metrics used to score the agreement with the labeled heads and largely used for big datasets such as MS COCO, appear to be less efficient when some heads are labeled in a more uncertain way as it was the case in several situations depending on the development stage, illumination conditions and head density. As a result, the weighted domain accuracy is proposed as a new metric [14]. The accuracy computed over image *i* belonging to the domain *d*, $AI_d(i)$, is classically defined as:

$$AI_d(i) = \frac{TP}{TP + FN + FP}$$

Where TP, FN and FP are respectively the number of true positive, false negative and false positive found in image $i$. The weighted domain accuracy (*WDA*) is the weighted average of all domain accuracies:

$$WDA = \frac{1}{D} \sum_{d=1}^{D} \frac{1}{n_d} * \sum_{i=1}^{n_d} AI_{di}$$

Where $D$ is the number of domains (sub-datasets) and $n_d$ is the number of images in domain d.

In a similar fashion to the first Global Wheat Challenge, the dataset will be split into three datasets: a training dataset containing all images, labels and metadata, a public validation dataset to be used be the challengers to evaluate their models, and a private test data set, used by he organizers of the challenge to score the performances on an"unseen" dataset. The training dataset contains from Europe (3657 images from 18 subdatasets), ,while the evaluation (1476 images from 11 subdatasets) and test (1373 images from 18 subdatasets) come from Africa, Asia, and North America sub-datasets. Note that the sub-dataset Usask_1 (Canada) has been moved from training dataset to public/ private dataset between GWC 2020 and GWC 2021. The exact content of public and private test set will be public at the end of the competition (4[th] July 2021).

5. **Conclusion**

The second edition of the Global Wheat Head Dataset, GWHD_2021, alongside the organization of a second Global Wheat Challenge is an important step for illustrating the usefulness of open and shared data across organizations to further improve high-throughput phenotyping methods. In comparison to the 2020 Kaggle competition, it represents 5 new countries, 22 new sub-datasets, 1200 new images and 120,000 new labeled wheat heads. Its revised organization and additional diversity are more representative of the type of images researchers and agronomists can acquire across the world. The revised metrics can help researchers to benchmark one-class localization models on a large range of acquisitions conditions. The competition is expected to accelerate the building of robust solutions thanks to its revised design. However, progress on representation of developing countries is still lacking and we are open to new contributions from South America, Africa and South Asia. We started to include nadir view photos from smartphones, to get a more comprehensive dataset and train reliable models for such affordable devices. Further, it is planned to release wheat head masks alongside the bounding box given the very large number of boxes that already exist and provide more associated metadata.

**Acknowledgments**

The work received support from ANRT for the CIFRE grant of Etienne David, co-funded by Arvalis for the project management.
The labelling work was supported by several companies and projects, including:
**Canada**: The Global Institute Food Security, University of Saskatchewan supported the organization of the competition.
**France**:  This work was supported by the French National Research Agency under the Investments for the Future Program, referred as ANR-16-CONV-0004 PIA #Digitag Institut Convergences Agriculture Numérique, Hiphen supported the organization of the competition.

**Japan**: Kubota supported the organization of the competition.

**Australia**: Grains Research and Development Corporation (UOQ2002-008RTX Machine learning applied to high-throughput feature extraction from imagery to map spatial variability and UOQ2003-011RTX INVITA - A technology and analytics platform for improving variety selection) supported competition.

We would like to thank the company "Human in the loop", which corrected and labeled the new datasets. The help of Frederic Venault (INRAe Avignon) was also precious to check the labelled images.